\documentclass[letterpaper, 10 pt, conference]{ieeeconf}  
\IEEEoverridecommandlockouts                              

\usepackage{00_ps}

\preprinttrue
\publishedtrue

\glsdisablehyper 
\DeclareAcronym{ap}{
    short = AP,
    long  = Action Propagation,
}
\DeclareAcronym{av}{
    short = AV,
    long  = Autonomous Vehicles,
}
\DeclareAcronym{c2c}{
    short = C2C,
    long  = Center-to-Center,
}
\DeclareAcronym{cav}{
    short = CAV,
    long  = Connected and Automated Vehicle,
}
\DeclareAcronym{cbf}{
    short = CBF,
    long  = Control Barrier Function,
}
\DeclareAcronym{cg}{
    short = CG,
    long = Center of Gravity,
    short-plural = s,
    long-plural-form = Centers of Gravity,
}
\DeclareAcronym{cnn}{
    short = CNN,
    long  = Convolutional Neural Network
}
\DeclareAcronym{cpm}{
    short = CPM,
    long  = Cyber-Physical Mobility
}
\DeclareAcronym{cpmlab}{
    short = CPM Lab,
    long  = Cyber-Physical Mobility Lab
}
\DeclareAcronym{dmpc}{
    short = DMPC,
    long  = distributed model predictive control
}
\DeclareAcronym{dql}{
    short = DQL,
    long  = Deep Q-Learning
}

\DeclareAcronym{hocbf}{
    short = HOCBF,
    long  = High-Order \ac{cbf},
}
\DeclareAcronym{il}{
    short = IL,
    long  = Imitation Learning,
    short-indefinite = an,
}
\DeclareAcronym{mappo}{
    short = MAPPO,
    long  = Multi-Agent \ac{ppo},
    short-indefinite = an,
}
\DeclareAcronym{maddpg}{
    short = MADDPG,
    long  = Multi-Agent Deep Deterministic Policy Gradient,
    short-indefinite = an,
}
\DeclareAcronym{mas}{
    short = MAS,
    long  = Multi-Agent System,
    short-indefinite = an,
}
\DeclareAcronym{mdp}{
    short = MDP,
    long  = Markov decision process,
    short-indefinite = an,
}
\DeclareAcronym{mg}{
    short = MG,
    long  = Markov Game,
    short-indefinite = an,
}
\DeclareAcronym{ml}{
    short = ML,
    long  = Machine Learning,
    short-indefinite = an,
}
\DeclareAcronym{mtv}{
    short = MTV,
    long  = Minimum Translation Vector,
    short-indefinite = an,
}
\DeclareAcronym{mpc}{
    short = MPC,
    long  = model predictive control,
    short-indefinite = an,
}
\DeclareAcronym{marl}{
    short = MARL,
    long  = Multi-Agent Reinforcement Learning,
    short-indefinite = an,
}
\DeclareAcronym{ocp}{
    short = OCP,
    long  = Optimal Control Problem,
    short-indefinite = an,
    long-indefinite = an,
}
\DeclareAcronym{per}{
    short = PER,
    long  = Prioritized Experience Replay
}
\DeclareAcronym{pomdp}{
    short = POMDP,
    long  = Partially Observable \ac{mdp}
}
\DeclareAcronym{pomg}{
    short = POMG,
    long  = Partially Observable \ac{mg}
}
\DeclareAcronym{ppo}{
    short = PPO,
    long  = Proximal Policy Optimization
}
\DeclareAcronym{qp}{
    short = QP,
    long  = Quadratic Program,
}
\DeclareAcronym{rhc}{
    short = RHC,
    long  = receding horizon control,
    short-indefinite = an,
}
\DeclareAcronym{rl}{
    short = RL,
    long  = Reinforcement Learning,
    short-indefinite = a,
}
\DeclareAcronym{sat}{
    short = SAT,
    long = Separating Axis Theorem,
    short-indefinite = an
}
\DeclareAcronym{som}{
    short = SOM,
    long  = Self Other-Modeling,
    short-indefinite = a,
}
\DeclareAcronym{ttcbf}{
    short = TTCBF,
    long  = Truncated Taylor \ac{cbf},
}
\DeclareAcronym{zsg}{
    short = ZSG,
    long  = Zero-Shot Generalization,
}
\providecommand{\doi}{10.1109/ITSC60802.2025.11423203} 

\makeatletter
\def\ps@IEEEtitlepagestyle{%
    \def\@oddfoot{\mycopyrightnotice}%
    \def\@evenfoot{}%
}
\makeatother

\ifpreprint
    \ifpublished
        \def\mycopyrightnotice{%
            \begin{minipage}{\textwidth}
                \centering \scriptsize
                \href{https://doi.org/\doi}{DOI: \doi} \copyright \the\year{} \copyrightOwnerFull{}. 
                Personal use of this material is permitted. Permission from \copyrightOwnerShort{} must be obtained for all other uses, in any current or future media, including reprinting/republishing this material for advertising or promotional purposes, creating new collective works, for resale or redistribution to servers or lists, or reuse of any copyrighted component of this work in other works.\hfill
            \end{minipage}
            \gdef\mycopyrightnotice{}
        }
    \else 
        \def\mycopyrightnotice{%
            \begin{minipage}{\textwidth}
                \centering \scriptsize
                This work has been submitted to the \copyrightOwnerFull{} for possible publication. Copyright may be transferred without notice, after which this version may no longer be accessible.
            \end{minipage}
            \gdef\mycopyrightnotice{}
        }
    \fi
\else
    \def\mycopyrightnotice{}
\fi

\def\BibTeX{{\rm B\kern-.05em{\sc i\kern-.025em b}\kern-.08em
  T\kern-.1667em\lower.7ex\hbox{E}\kern-.125emX}}

\begin{document}
\title{\LARGE \bf
    A Real-Time Control Barrier Function-Based Safety Filter for Motion Planning with Arbitrary Road Boundary Constraints
    \thanks{This research was supported by the Bundesministerium für Digitales und Verkehr (German Federal Ministry for Digital and Transport) within the project ``Harmonizing Mobility'' (grant number 19FS2035A).}
}

\author{
    Jianye Xu$^{1}$\,\orcidlink{0009-0001-0150-2147},~\IEEEmembership{Student~Member,~IEEE},
    Chang Che$^{1}$\,\orcidlink{0009-0001-0577-8845},
    Bassam Alrifaee$^{2}$\,\orcidlink{0000-0002-5982-021X},~\IEEEmembership{Senior Member, ~IEEE}
    \thanks{$^{1}$Department of Computer Science, RWTH Aachen University, Germany, \texttt{\{last name\}@embedded.rwth-aachen.de}}
    \thanks{$^{2}$Department of Aerospace Engineering, University of the Bundeswehr Munich, Germany, \texttt{bassam.alrifaee@unibw.de}}
}
    \maketitle
\thispagestyle{IEEEtitlepagestyle}
\begin{abstract}
\noindent
We present a real-time safety filter for motion planning, including those that are learning-based, using Control Barrier Functions (CBFs) to provide formal guarantees for collision avoidance with road boundaries. A key feature of our approach is its ability to directly incorporate road geometries of arbitrary shape that are represented as polylines without resorting to conservative overapproximations. We formulate the safety filter as a constrained optimization problem as a Quadratic Program (QP), which achieves safety by making minimal, necessary adjustments to the control actions issued by the nominal motion planner. We validate our safety filter through extensive numerical experiments across a variety of traffic scenarios featuring complex road boundaries. The results confirm its reliable safety and high computational efficiency (execution frequency up to 40 Hz). Code reproducing our experimental results and a video demonstration are available at \href{https://github.com/bassamlab/SigmaRL}{\small github.com/bassamlab/SigmaRL}.
\end{abstract}

\section{Introduction} \label{sec:introduction}
\Acp{av} hold the promise of transforming transportation by enhancing safety and efficiency \cite{fagnant2015preparing}. A cornerstone enabling \acp{av} is motion planning, which involves computing safe and feasible trajectories from the vehicle's current state to a desired goal while respecting vehicle dynamics and environmental constraints \cite{paden2016survey}. In this work, we focus specifically on collision avoidance with road boundaries in motion planning.

Motion planning methods for \acp{av} can be broadly categorized based on their underlying principles, including optimization-, sampling-, search-, and learning-based methods. Optimization-based methods formulate motion planning as an optimization problem, typically minimizing a cost function subject to constraints that include vehicle dynamics and collision avoidance \cite{kirk2004optimal}. Handling collision avoidance with arbitrarily shaped road boundaries within this framework often requires incorporating non-convex constraints, which can be computationally demanding to solve in real-time \cite{zhang2021optimizationbased}. Common strategies involve approximating these constraints through techniques like linearization, convex restriction \cite{scheffe2023sequential}, or convex relaxation \cite{alrifaee2014centralized}. While these approximations can improve computational tractability, they may introduce conservatism, limiting the vehicle's maneuverability \cite{xu2025learningbased}, or make it difficult to quantify the approximation error \cite{schulman2014motion}. Sampling-based methods, such as the Rapidly Exploring Random Tree (RRT) \cite{tuncali2019rapidlyexploring}, explore the state space by sampling configurations and connecting them to build a collision-free path. Graph-based methods, like A* search \cite{hart1968formal}, discretize the configuration space and search for an optimal path. While effective, ensuring safe and efficient motion planning involving complex road boundary constraints can require dense sampling or fine discretization, potentially increasing computational costs. In recent years, learning-based methods such as imitation learning \cite{pomerleau1988alvinn} and \ac{rl} \cite{shalev-shwartz2016safe, du2023reinforcement} have shown promising performance in learning complex motion planning policies. However, a remaining challenge is the difficulty in providing formal safety guarantees \cite{koopman2017autonomous}, especially in scenarios not encountered during training. This challenge motivates the development of complementary safety certification approaches.

\acp{cbf}, grounded in control theory, provide a formal framework for ensuring the forward invariance of a designated safe set for a dynamical system \cite{ames2017control}. When applied to collision avoidance with road boundaries, the safe set typically represents the subset of the state space in which the vehicle remains within the drivable set. \acp{cbf} can be used to certify \textit{a posteriori} whether a control action issued by a (potentially unsafe) motion planner remains within this safe set \cite{wabersich2023datadriven}. A key advantage of \acp{cbf} is their ability to be embedded as affine constraints in a \ac{qp}, which makes them more computationally tractable than other formal safety certification tools such as reachability analysis \cite{pek2021failsafe, scheffe2024limiting}. For instance, concerning the safety requirement of \acp{av}, \cite{liufu2025collaborative} constructed a nonconvex \ac{qp} with \ac{cbf} constraints that can be effectively solved with collaborative neural dynamics in real time, achieving a breakthrough in dealing with nonconvex control problems.

However, applying \acp{cbf} directly to enforce constraints imposed by arbitrary road boundaries presents a significant challenge: the construction of an appropriate \ac{cbf}. Standard \ac{cbf} formulations typically require the function defining the boundary of the safe set, such as a function representing the distance to the road boundaries, to be smooth and continuously differentiable \cite{ames2017control}. In practice, however, road geometries are often specified in nonanalytic forms, such as polylines. Constructing a smooth and differentiable barrier function from such representations is generally nontrivial and often infeasible \cite{chen2025control}. As a result, prior applications of \acp{cbf} in motion planning for \acp{av} commonly avoid scenarios that involve general road boundary constraints. Instead, they focus on settings where the safe set can be described using simple geometric approximations. These include longitudinal control tasks such as adaptive cruise control \cite{ames2014control, ames2017control}, structured scenarios like highway merging under idealized assumptions \cite{xiao2021decentralized, xiao2021bridging}, or environments characterized by fixed-width roads or curvature-invariant lanes \cite{ames2019control, khajenejad2021tractable, zheng2024barrierenhanced}. 

We overcome the above limitation by developing a \ac{cbf}–based safety filter that enables real-time safety certification of motion planning on arbitrary road boundaries presented as polylines. Our safety filter directly considers the road boundaries without overapproximating them while maintaining real-time performance (up to \SI{40}{\hertz} execution frequency). To our knowledge, no previous \ac{cbf}-based safety filter jointly achieves these properties. Extensive numerical experiments on roads with complex road boundaries confirm its computational efficiency and effectiveness.

\section{Problem Statement} \label{sec:problem}
Given a \emph{nominal} motion planner, such as a learning-based one, that controls a vehicle in a two-dimensional environment bounded by a left and a right road boundary, we aim to certify its safety. At time $t$, the nominal planner outputs the control action $\bm{u}_{\mathrm{nom}}(t)$ for the driving task (e.g., path following). Our objective is to ensure that the vehicle does not collide with the road boundaries during an interval $[t_0,t_f]$, where $t_0$ denotes the initial time of safety certification, and the final time $t_f$ can be infinite. If the nominal control would lead to a collision, we compute a \emph{corrective} control action $\bm{u}(t)$ that minimally deviates from $\bm{u}_{\mathrm{nom}}(t)$.

\begin{figure}
    \centering
    \includegraphics[width=1.0\linewidth]{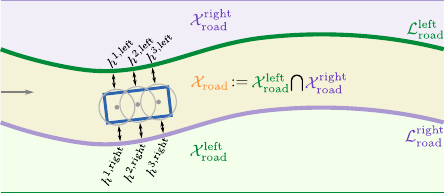}
    \caption{Conceptual visualization of drivable sets: 
    $\mathcal{X}^{\mathrm{left}}_{\mathrm{road}}$ in green (beneath the left boundary $\mathcal{L}^{\mathrm{left}}_{\mathrm{road}}$), 
    $\mathcal{X}^{\mathrm{right}}_{\mathrm{road}}$ in purple (above the right boundary $\mathcal{L}^{\mathrm{right}}_{\mathrm{road}}$), and 
    $\mathcal{X}_{\mathrm{road}}$ in orange (within both boundaries, jointly imposed by them). 
    Vehicle in blue, circles approximating it in gray, and \acp{cbf} $h$ conceptually depicted in black arrows (details in \cref{sec:vehicle-approximation}).}
    \label{fig_drivable_sets}
\end{figure}

Consider the vehicle modeled by a nonlinear input-affine control system
\begin{equation} \label{eq:dynamics}
    \dot{\bm{x}} = f\bigl(\bm{x}\bigr) + g(\bm{x})\bm{u},
\end{equation}
where $\bm{x} \in \mathcal{X}\subset\mathbb{R}^n$ is the state and $\bm{u} \in \mathcal{U}\subset\mathbb{R}^m$ is the control input. We assume that the left and right road boundaries of the vehicle at time $t$ can have arbitrary shapes. Each boundary is represented as a polyline, i.e., a set of connected line segments defined by discrete points that can be obtained from perception data or high-definition maps. We denote each polyline by $\mathcal{L}_{\mathrm{road}}^{i}(t), \forall i \in \{ \mathrm{left,right} \}$. Let the drivable set imposed by boundary $i$ be defined as
\begin{equation}
    \mathcal{X}_{\mathrm{road}}^{i}(t)
    =
    \Bigl\{
        \bm{x}\in\mathcal{X}
        \mid
        \mathrm{sd}\bigl(\mathcal{O}_{\mathrm{veh}}(\bm{x}),\mathcal{L}_{\mathrm{road}}^{i}(t)\bigr) \ge 0
    \Bigr\},
\end{equation}
where $\mathcal{O}_{\mathrm{veh}}(\bm{x})\subset\mathbb{R}^{2}$ is the geometric occupancy of the vehicle, and $\mathrm{sd}(\cdot,\cdot)$ is a signed-distance function that is positive when the vehicle occupancy lies inside the boundary. The overall drivable set is the intersection
\begin{equation}\label{eq:drivable-set}
    \mathcal{X}_{\mathrm{road}}(t)
    \coloneqq
    \mathcal{X}_{\mathrm{road}}^{\mathrm{left}}(t) \cap \mathcal{X}_{\mathrm{road}}^{\mathrm{right}}(t).
\end{equation}
Because no geometric overapproximation is applied, $\mathcal{X}_{\mathrm{road}}(t)$ presents the true drivable set. Safety requires
\begin{equation}\label{eq:collision-avoidance-general}
    \bm{x}(t)\in\mathcal{X}_{\mathrm{road}}(t), \quad \forall t\in[t_0,t_f].
\end{equation}
\Cref{fig_drivable_sets} depicts an example visualizing the left and right road boundaries, $\mathcal{L}^{\mathrm{left}}_{\mathrm{road}}$ and $\mathcal{L}^{\mathrm{right}}_{\mathrm{road}}$, and the imposed drivable sets, $\mathcal{X}^{\mathrm{left}}_{\mathrm{road}}, \mathcal{X}^{\mathrm{right}}_{\mathrm{road}}$, and $\mathcal{X}_{\mathrm{road}}$.

Directly solving \iac{ocp} containing \eqref{eq:collision-avoidance-general} is computationally demanding due to its nonconvex nature. The \ac{cbf}-based approach addresses this problem by formulating a computationally efficient \ac{qp} problem, commonly denoted as the \ac{cbf}-\ac{qp} framework \cite{ames2019control}.

\subsection{Control Barrier Function Preliminaries}
A set $\mathcal{C} \subset \mathcal{X}$ is called \emph{forward invariant} for system \eqref{eq:dynamics} if any trajectory starting in $\mathcal{C}$ at time $t_0$ remains in $\mathcal{C}$ for all $t \ge t_0$ \cite{ames2019control}. A function $\alpha: \mathbb{R} \to \mathbb{R}$ is of \emph{extended class $\mathcal{K}$} if it is Lipschitz continuous, strictly increasing, and satisfies $\alpha(0)=0$ \cite{ames2019control}. The \emph{relative degree} $r \in \mathbb{N}$ of a continuously differentiable function $h: \mathcal{X} \to \mathbb{R}$ w.r.t. system \eqref{eq:dynamics} is the number of times we must differentiate it along the system dynamics until the control input $\bm{u}$ appears \cite{xiao2019control}.

Collision-avoidance constraints often exhibit relative degrees higher than one. For example, when the control input is acceleration, a commonly used distance-based constraint that is defined over position must be differentiated twice before the control input appears. This necessitates \acp{hocbf} \cite{xiao2019control}. For each road boundary $i \in \{\mathrm{left,right}\}$, let $h^i(\bm{x},t)$ be a candidate \ac{cbf} of relative degree $r^i \in \mathbb{N}$ that defines a safe set $C^i(t) \coloneqq \{ \bm{x} \in \mathcal{X} \mid h^i(\bm{x},t) \ge 0 \} \subseteq \mathcal{X}_{\mathrm{road}}^i(t)$. Instead of using the standard \ac{hocbf} proposed in \cite{xiao2019control}, which requires tuning multiple class $\mathcal{K}$ functions, we apply our \ac{ttcbf} proposed in \cite{xu2025highorder}, which requires only a single class $\mathcal{K}$ function. Given a sampling period $\Delta t \in \mathbb{R}$ in the discrete-time domain, it enforces collision-avoidance constraints as
\begin{equation} \label{eq:ttcbf}
    \Delta t \dot{h}^i + \cdots + \frac{1}{r^i!}\Delta t^{r^i} {h^{i}}^{(r^i)} + \alpha\bigl(h^i\bigr) \ge \gamma \Delta t^{r^i+1},
\end{equation}
where ${h^i}^{(r^i)}$ denotes the $r^i$th time derivative of $h^i$ (which invokes the control input), and $\gamma \in \mathbb{R}$ is a tuning parameter. As shown in \cite{xu2025highorder}, \eqref{eq:ttcbf} renders set $C^i$ forward invariant, i.e., $h^i \ge 0$ for all $t \ge t_0$.

\subsection{Optimal Control Problem Formulation}
We formulate \iac{ocp} for the safety filter using the \ac{cbf}-\ac{qp} framework, which will be solved at each time step $k$.
\begin{subequations} \label{eq:ocp-original}
\begin{align}
    \min_{\bm{u}_k}\quad & \bigl\lVert \bm{u}_k - \bm{u}_{\mathrm{nom},k}\bigr\rVert_R^2 \label{eq:ocp-objective-function} \\
    \text{s.t.}\quad 
    & \text{Constraint } \eqref{eq:ttcbf}, \forall i \in \{\mathrm{left,right}\}, \label{eq:ocp:collision-avoidance} \\
    & \bm{u}_{\min} \le \bm{u}_k \le \bm{u}_{\max}. \label{eq:ocp-u-max-min}
\end{align}
\end{subequations}
Here, $R \in \mathbb{R}^{m \times m}$ is a positive-definite weighting matrix usually chosen to be diagonal, and $\bm{u}_{\min}$ and $\bm{u}_{\max}$ are the control bounds. Since \eqref{eq:ocp:collision-avoidance} renders $\mathcal{C}_i$ forward invariant $\forall i \in \{\mathrm{left,right}\}$, it holds that $\bm{x} \in C^{\mathrm{left}}(t) \cap C^{\mathrm{right}}(t) \subseteq \mathcal{X}_{\mathrm{road}}(t), \forall t \ge t_0$. Therefore, the vehicle state stays within the actual drivable set $\mathcal{X}_{\mathrm{road}}$ for all $t \ge t_0$. The remaining task is to construct \acp{cbf} $h^i(\bm{x},t)$ such that $C^{\mathrm{left}}(t) \cap C^{\mathrm{right}}(t)$ tightly approximates $\mathcal{X}_{\mathrm{road}}(t)$ without sacrificing real-time solvability. 

\section{\ac{cbf}-Based Safety Filter}
We present our \ac{cbf}-based safety filter that adapts a distance function to tightly approximate the drivable set defined in \eqref{eq:drivable-set}, without introducing conservative overapproximations of the road boundaries.

We describe the distance function in \cref{sec:pseudo-distance} and how we adopt it to construct \acp{cbf} in \cref{sec:cbf-construction}. In \cref{sec:vehicle-approximation}, we present how we approximate the geometric occupancy of the vehicle with a set of circles. Finally, in \cref{sec:case-study}, we conduct a case study showcasing how to apply our approach to the well-known nonlinear kinematic bicycle model. 

\subsection{Pseudo-Distance} \label{sec:pseudo-distance} 
The pseudo-distance was first introduced in \cite{ziegler2014trajectory} to ensure the continuity and differentiability of the distance-to-road-boundary constraint in Newton-type optimization problems, with road boundaries represented as polylines. Therefore, this pseudo-distance can be an ideal candidate \ac{cbf} to impose safety distances to road boundaries presented as polylines.

\begin{figure}[t!] 
    \centering
    \includegraphics[width=1\linewidth]{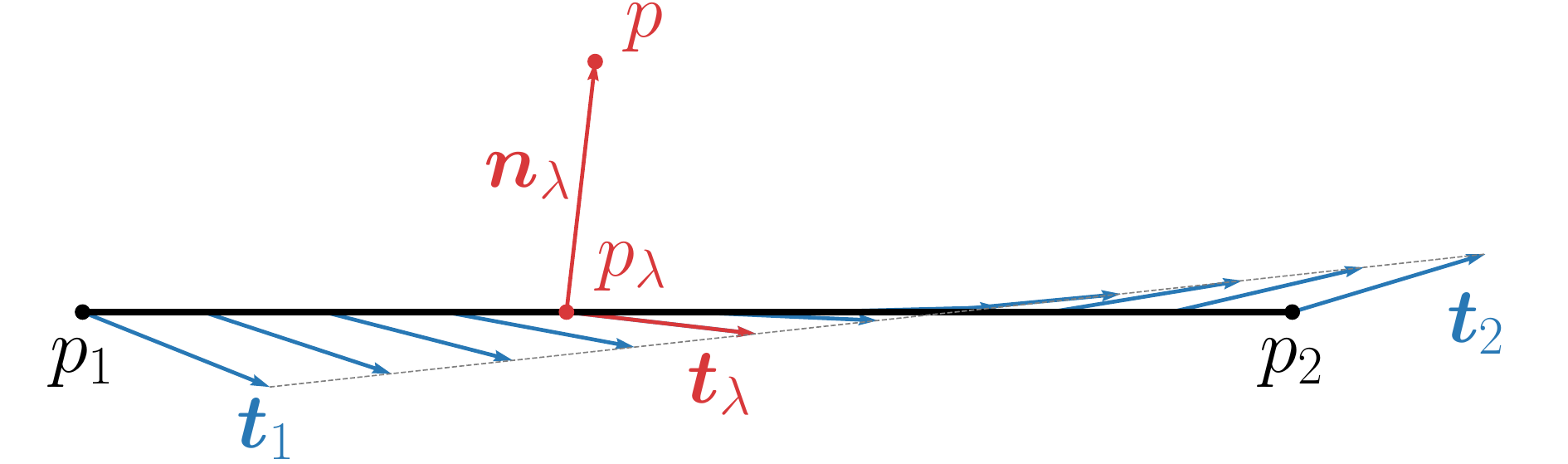}
    \caption{Pseudo-distance from an arbitrary point $p$ to the line segment $G=(\overline{p_{1}p_{2}},\bm{t}_{1},\bm{t}_{2})$ \cite{ziegler2014trajectory}.}
    \label{fig_pseudo_distance_segment}
\end{figure}

\begin{figure}[t!]
    \centering
    \includegraphics[width=1\linewidth]{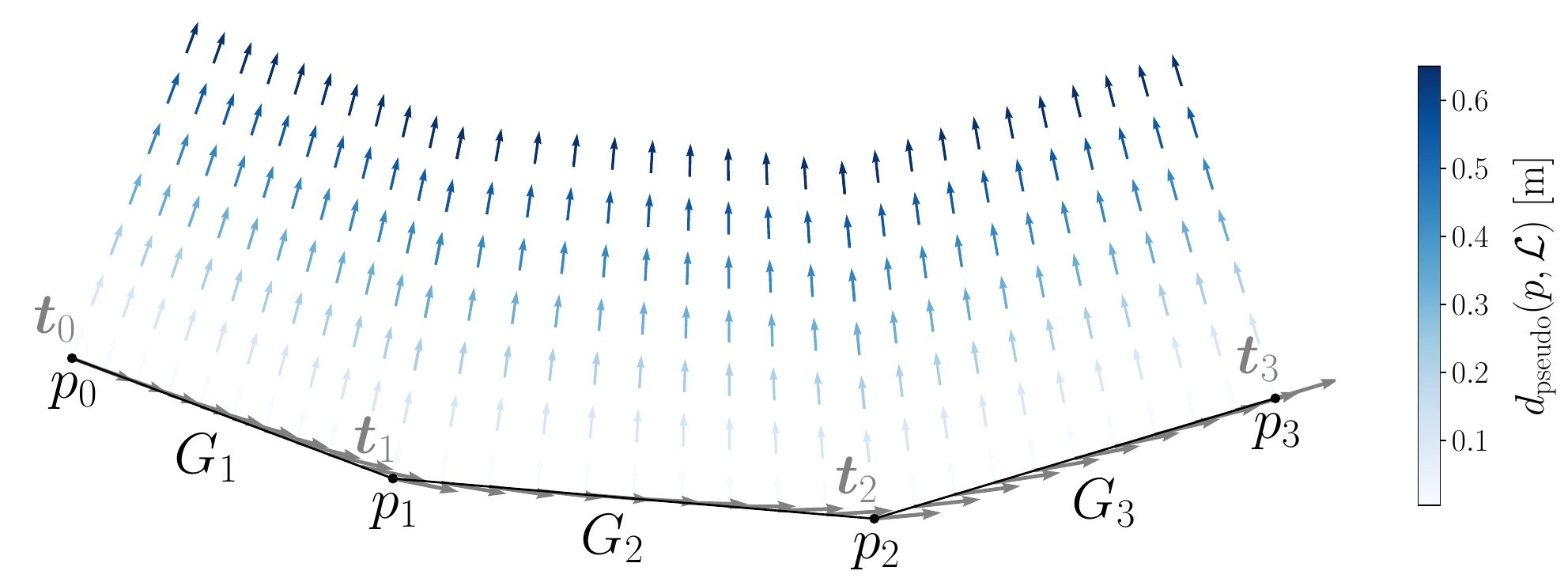}
    \caption{Gradient field of the pseudo-distance from point $p$ to an example polyline $\mathcal{L}$ with three line segments $\{G_{j}\}_{j=1}^{3}$.}
    \label{fig_pseudo_distance_polyline}
\end{figure}

First, we define a single line segment of a polyline as a tuple $G=(\overline{p_{1}p_{2}},\bm{t}_{1},\bm{t}_{2})$, whose end-points $p_{1}, p_{2} \in \mathbb{R}^2$ carry predefined tangent vectors $\bm{t}_{1}, \bm{t}_{2} \in \mathbb{R}^2$. One simple way to define a tangent vector is to use a (possibly normalized) vector defined by neighboring points, e.g., $\bm{t}_1 = \overline{p_{0}p_{2}}$. By linearly interpolating along the segment, a point $p_{\lambda}$ on the segment and its pseudo tangent vector $\bm{t}_{\lambda}$ can be given as
\begin{equation*}
    \begin{aligned}
    p_{\lambda}&=\lambda p_{2} + (1-\lambda)p_{1},\\
    \bm{t}_{\lambda}&=\lambda \bm{t}_{2} + (1-\lambda)\bm{t}_{1},
    \end{aligned}
\end{equation*}
where $\lambda \in [0, 1]$ is the interpolation parameter. The magnitude of the pseudo-distance from an arbitrary external point $p$ to this line segment $G$ is defined as the norm of the vector $\bm{n}_{\lambda} \coloneqq p - p_{\lambda}$ that is orthogonal to $\bm{t}_{\lambda}$, i.e.,
\begin{equation}
    d_{\mathrm{pseudo}}(p,G) = \|\bm{n}_{\lambda}\|\quad \text{s.t.}\; \bm{n}_{\lambda}^{\top}\bm{t}_{\lambda}=0.
\end{equation}
\Cref{fig_pseudo_distance_segment} illustrates this computation principle. 
For a polyline $\mathcal{L}$ consisting of $n_p$ line segments $\{G_{j}\}_{j=1}^{n_p}$, the pseudo-distance from a point $p$ to the polyline is defined as the minimum pseudo-distance from $p$ to any of the segments
\begin{equation} \label{eq:pseudo-distance-point-polyline}
   d_{\mathrm{pseudo}}(p,\mathcal{L}) \coloneqq \min_{1 \leq j \leq n_p} d_{\mathrm{pseudo}}(p, G_j).
\end{equation}
In \Cref{fig_pseudo_distance_polyline}, the colored arrows visualize the pseudo-distance gradient field corresponding to an example polyline consisting of three line segments, with the direction of the distance indicated by the arrow orientation and the magnitude encoded by color. While the Euclidean distance to a polyline typically exhibits a discontinuous gradient near the junctions of adjacent segments, the pseudo-distance ensures gradient continuity by interpolating tangent vectors along the segments, thus enabling a smooth transition of the distance gradient across the junctions of segments \cite{ziegler2014trajectory}.

\subsection{\ac{cbf} Construction} \label{sec:cbf-construction}
The pseudo‐distance computes the distance between a point $p$ and a polyline $\mathcal{L}$. To apply it between the vehicle and the road boundary $\mathcal{L}_{\mathrm{road}}^i$, $i\in\{\mathrm{left},\mathrm{right}\}$, we approximate the vehicle's occupancy $\mathcal{O}_{\mathrm{veh}}(\bm{x})$ by a set of $n_{\mathrm{cir}}$ circular occupancies $\bigl\{\mathcal{O}^j_{\mathrm{cir}}(\bm{x})\bigr\}_{j=1}^{n_{\mathrm{cir}}}$, yielding 
\begin{equation} \label{eq:circles-overapproximate}
  \bigcup_{j=1}^{n_{\mathrm{cir}}}\mathcal{O}^j_{\mathrm{cir}}(\bm{x})
  \;\supseteq \; 
  \mathcal{O}_{\mathrm{veh}}(\bm{x}).
\end{equation}
Each $\mathcal{O}^j_{\mathrm{cir}}(\bm{x})$ represents a circle with center $\bm{c}^j_{\mathrm{cir}}(\bm{x})$ and radius $r^j_{\mathrm{cir}}(\bm{x})$. We construct a signed-distance function which we employ as our \ac{cbf} $h^{j,i}$ to control the distance between each circle $j \in \{1,\ldots,n_{\mathrm{cir}}\}$ and each road boundary $i \in \{\mathrm{left,right}\}$ as
\begin{equation} \label{eq:new-cbf}
\begin{aligned}
    h^{j,i}(\bm{x}, t) &\coloneqq \mathrm{sd}\bigl(\mathcal{O}^j_{\mathrm{cir}}(\bm{x}), \mathcal{L}_{\mathrm{road}}^i(t)\bigr) \\
    &\coloneqq d_{\mathrm{pseudo}}\bigl(\bm{c}^j_{\mathrm{cir}}(\bm{x}), \mathcal{L}_{\mathrm{road}}^i(t)\bigr) - r^j_{\mathrm{cir}}(\bm{x}),
\end{aligned}
\end{equation}
where $d_{\mathrm{pseudo}}(\cdot,\cdot)$ computes the pseudo‐distance from a point to a polyline as in \eqref{eq:pseudo-distance-point-polyline}. If $h^{j,i} \ge 0, \forall j \in \{1,\ldots,n_\mathrm{cir}\}$, all circles do not collide with the road boundary $i$; therefore, the vehicle also does not collide with it due to the overapproximation \eqref{eq:circles-overapproximate}. 

We reformulate the collision-avoidance constraint \eqref{eq:ocp:collision-avoidance} as 
\begin{equation} \label{eq:ttcbf-reformulated}
\begin{aligned}
    &\Delta t \dot{h}^{j,i}  + \cdots + \frac{1}{r^{j,i}!}\Delta t^{r^{j,i}} {h^{j,i}}^{(r^{j,i})} + \alpha\bigl(h^{j,i}\bigr) \\ 
    & \ge \gamma \Delta t^{r^{j,i}+1}, \forall j \in \{1,\ldots,n_{\mathrm{cir}}\}, \forall i \in \{\mathrm{left,right}\}.
\end{aligned}
\end{equation}
Note that \eqref{eq:ttcbf-reformulated} represents a set of $2n_{\mathrm{cir}}$ collision-avoidance constraints. \Cref{fig_drivable_sets} depicts an example with $n_{\mathrm{cir}}=3$, resulting in six \acp{cbf}, $h^{1,\mathrm{left}},h^{2,\mathrm{left}},h^{3,\mathrm{left}},h^{1,\mathrm{right}},h^{2,\mathrm{right}}$, and $h^{3,\mathrm{right}}$. 

\subsection{Vehicle Geometry Approximation} \label{sec:vehicle-approximation}
\begin{figure}[t!]
    \centering
    \includegraphics[width=0.5\linewidth]{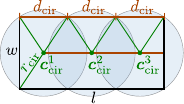}
    \caption{Overapproximation of a rectangular vehicle with length $\ell$ and width $w$ using three identical, equidistant circles with radius $r_{\mathrm{cir}}$.}
    \label{fig_rectangle_approximation_illustration}
\end{figure}

For a given number of $n_{\mathrm{cir}}$ circles, we aim to determine the minimum radius $r_{\mathrm{cir}}$ required to fully cover the occupancy of the vehicle for a tight approximation. For simplicity, we assume the vehicle to be rectangular with width $w$ and length $\ell$, and circles are identical and uniformly distributed along the longitudinal axis of the rectangle. Let $d_{\mathrm{cir}}$ denote the distance between the centers of adjacent circles. For complete coverage, the circles must cover the four edges of the rectangle. The optimal configuration places the first and last circle centers such that the two width edges are just covered, and the radius is minimized while the length edges are just covered. This leads to two conditions, $n_{\mathrm{cir}} d_{\mathrm{cir}} = \ell$ and $d_{\mathrm{cir}}^2 + w^2 = (2r_{\mathrm{cir}})^2$. Substituting $d_{\mathrm{cir}} = \ell / n_{\mathrm{cir}}$ in the second condition and solving for $r_{\mathrm{cir}}$ yields
\begin{equation} \label{eq:minimum-radius}
    r_{\mathrm{cir}} = \sqrt{\left(\frac{\ell}{2n_{\mathrm{cir}}}\right)^2 + \left(\frac{w}{2}\right)^2}.
\end{equation}
\Cref{fig_rectangle_approximation_illustration} illustrates this principle with three circles.

\subsection{Case Study with Kinematic Bicycle Model} \label{sec:case-study}
\begin{figure}[t!]
    \centering
    \includegraphics[width=0.8\linewidth]{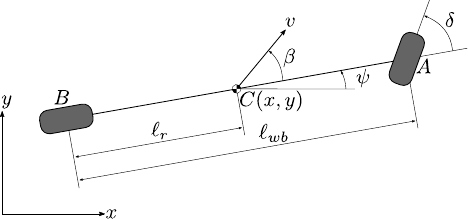}
    \caption{
        The kinematic bicycle model.
        $C$: center of gravity;
        $x, y$: $x$- and $y$-coordinates;
        $v$: velocity;
        $\beta$: slip angle;
        $\psi$: yaw angle;
        $\delta$: steering angle;
        $\ell_{wb}$: wheelbase;
        $\ell_{r}$: rear wheelbase.
    }
    \label{fig:bicycle-model}
\end{figure}

In this section, we conduct a case study deriving the concrete expression of the collision-avoidance constraint \eqref{eq:ttcbf-reformulated} for the nonlinear kinematic bicycle model \cite{rajamani2006vehicle}. This model captures the essential dynamics required for motion planning and control of \acp{av} and is adequate for scenarios involving moderate acceleration \cite{polack2017kinematic}. It approximates the vehicle as a single-track model with two wheels, as depicted in \cref{fig:bicycle-model}.

The state vector $\bm{x} \coloneqq [x, y, \psi, v, \delta]^\top \in \mathbb{R}^5$ contains the position $x, y$, heading $\psi$ (also called yaw or rotation), speed $v$, and steering angle $\delta$ (in the global coordinate system). The control input vector $\bm{u} \coloneqq [u_v, u_\delta]^\top \in \mathbb{R}^2$ consists of acceleration $u_v$ and steering rate $u_\delta$. The dynamics of the kinematic bicycle model are given by
\begin{equation} \label{eq:kinematic-bicycle-model}
    \dot{\bm{x}} =
    \begin{bmatrix}
        v \cos(\psi + \beta) \\
        v \sin(\psi + \beta) \\
        \dfrac{v}{\ell_{wb}} \tan(\delta) \cos(\beta) \\
        0 \\
        0
    \end{bmatrix}
    +
    \begin{bmatrix}
    0 & 0 \\
    0 & 0 \\
    0 & 0 \\
    1 & 0 \\
    0 & 1
    \end{bmatrix}
    \begin{bmatrix}
        u_v \\
        u_\delta
    \end{bmatrix},
\end{equation}
where $\ell_{wb} \in \mathbb{R}$ denotes the wheelbase of the vehicle, and the slip angle $\beta = \tan^{-1} \bigl( \dfrac{\ell_r}{\ell_{wb}} \tan \delta \bigr)$, with $\ell_r \in \mathbb{R}$ denoting the rear wheelbase.

To derive the expression of \eqref{eq:ttcbf-reformulated}, we need to compute the time derivatives of $h^{j,i}$ for up to $r^{j,i}$th order, i.e., $\dot{h}^{j,i},\ldots,{h^{j,i}}^{(r^{j,i})}$. Since our \ac{cbf} $h^{j,i}$ \eqref{eq:new-cbf} is a distance-related function, we need to differentiate it twice until the control input (acceleration and steering rate) appears. Therefore, $r_{j,i}=2$, simplifying \eqref{eq:ttcbf-reformulated} to 
\begin{equation} \label{eq:ttcbf-rereformulated}
\begin{aligned}
    \Delta t \dot{h}^{j,i} + \frac{1}{2} & \Delta t^2 \ddot{h}^{j,i} + \alpha(h^{j,i}) \ge \gamma \Delta t^3, \\ &\forall j \in \{1,\ldots,n_{\mathrm{cir}}\}, \forall i \in \{\mathrm{left,right}\}.
\end{aligned}
\end{equation}
To compute $\dot{h}^{j,i}$ and $\ddot{h}^{j,i}$, we consider the vehicle-fixed coordinate with the origin being the geometric center of the vehicle and the $x$-axis aligning with its longitudinal axis. Let the leftmost circle have index one, and the rightmost $n_{\mathrm{cir}}$. Given the distance $d_{\mathrm{cir}}=\ell/n_{\mathrm{cir}}$ between adjacent circles, we easily obtain the position of circle $j$'s center $\bm{c}_{\mathrm{cir}}^{\underline{j}} \in \mathbb{R}^{2}, \forall j \in \{ 1,\ldots,n_{\mathrm{cir}} \}$, in the vehicle-fixed coordinate as (underscore indicates vehicle-fixed coordinate)
\begin{equation} \label{eq:circle-position-vehicle-fixed-coordinate}
    \bm{c}_{\mathrm{cir}}^{\underline{j}} \coloneqq
    \begin{bmatrix}
        \bm{c}_{\mathrm{cir},x}^{\underline{j}} \\
        \bm{c}_{\mathrm{cir},y}^{\underline{j}}
    \end{bmatrix}
    = 
    \begin{bmatrix}
        \bigl(-\frac{1}{2} + \frac{2j-1}{2n_{\mathrm{cir}}}\bigr)\ell \\
        0
    \end{bmatrix}.
\end{equation}
Applying coordinate transformation (rotation by the vehicle's heading $\psi$ and translating by its position $x, y$) yields circle $j$'s position in the global coordinate as
\begin{equation} \label{eq:circle-center}
    \bm{c}^j_{\mathrm{cir}}
    = 
    \begin{bmatrix}
        \bm{c}^j_{\mathrm{cir},x} \\
        \bm{c}^j_{\mathrm{cir},y}
    \end{bmatrix}
    =
    \begin{bmatrix}
        \cos\psi & -\sin\psi \\
        \sin\psi & \cos\psi
    \end{bmatrix}
    \bm{c}_{\mathrm{cir}}^{\underline{j}}
    +
    \begin{bmatrix}
        x \\
        y
    \end{bmatrix}.
\end{equation}
Recall that $h^{j,i}$ \eqref{eq:new-cbf} is a function computing the pseudo-distance from circle $j$'s center to the polyline $i$ minus circle $j$'s radius $r_{\mathrm{cir}}^j$. The time derivatives of $h^{j,i}$ depend only on circle $j$'s position for the following three reasons: 1) circle $j$'s rotation does not affect $h^{j,i}$, 2) its radius $r_{\mathrm{cir}}^j$ is time invariant, and 3) we assume road boundaries are also time invariant (which is a mild assumption since road boundaries are generally static). Therefore, $\dot{h}^{j,i} = \nabla {h^{j,i}}^\top \dot{\bm{c}}^j_{\mathrm{cir}}$, where $\nabla h^{j,i} \coloneqq \bigl[\frac{\partial h^{j,i}}{\partial \bm{c}_{\mathrm{cir},x}}, \frac{\partial h^{j,i}}{\partial \bm{c}_{\mathrm{cir},y}} \bigr]^\top \in \mathbb{R}^{2}$ denotes the gradient vector of $h^{j,i}$. To compute $\dot{\bm{c}}^j_{\mathrm{cir}} \coloneqq \bigl[ \dot{\bm{c}}^j_{\mathrm{cir},x}, \dot{\bm{c}}^j_{\mathrm{cir},y} \bigr]^\top \in \mathbb{R}^{2}$, we take the time derivative of \eqref{eq:circle-center} and obtain (noticing $\bm{c}^{\underline{j}}_{\mathrm{cir},x}$ is time invariant and $\bm{c}^{\underline{j}}_{\mathrm{cir},y}=0$, see \eqref{eq:circle-position-vehicle-fixed-coordinate})
\begin{equation} \label{eq:dt-circle}
    \begin{aligned}
        \dot{\bm{c}}^j_{\mathrm{cir},x} &= -\sin\psi \; \dot{\psi} \; \bm{c}^{\underline{j}}_{\mathrm{cir},x} + \dot{x}, \\
        \dot{\bm{c}}^j_{\mathrm{cir},y} &= \cos\psi \; \dot{\psi} \; \bm{c}^{\underline{j}}_{\mathrm{cir},x} + \dot{y}.
    \end{aligned}
\end{equation}
The second time derivative of $h^{j,i}$ is given by $\ddot{h}^{j,i} = \nabla {h^{j,i}}^\top  \ddot{{\bm{c}}}^j_{\mathrm{cir}} + \dot{{\bm{c}}}^{j\;\top}_{\mathrm{cir}} H \; \dot{{\bm{c}}}^{j}_{\mathrm{cir}}$, where $H \in \mathbb{R}^{2 \times 2}$ denotes the Hessian matrix of $h^{j,i}$.
Taking time derivative of \eqref{eq:dt-circle} yields
\begin{equation} \label{eq:ddt-circle}
\begin{aligned}
    \ddot{\bm{c}}^j_{\mathrm{cir},x} &= -\cos\psi \; \dot{\psi}^2 \; \bm{c}^{\underline{j}}_{\mathrm{cir},x} - \sin\psi \; \ddot{\psi} \; \bm{c}^{\underline{j}}_{\mathrm{cir},x} + \ddot{x} \notag \\
    \ddot{\bm{c}}^j_{\mathrm{cir},y} &= -\sin\psi \; \dot{\psi}^2 \; \bm{c}^{\underline{j}}_{\mathrm{cir},x} + \cos\psi \; \ddot{\psi} \; \bm{c}^{\underline{j}}_{\mathrm{cir},x} + \ddot{y} \notag \\
\end{aligned}
\end{equation}
The first time derivatives $\dot{x}, \dot{y}, \dot{\psi}$ are directly given by \eqref{eq:kinematic-bicycle-model}, and taking another time derivative of them easily yields their second time derivatives $\ddot{x},\ddot{y},\ddot{\psi}$, which we omit due to space limitations. Furthermore, the gradient vector $\nabla h^{j,i}$ and Hessian matrix $H$ of $h^{j,i}$ can be easily obtained using numerical differentiation \cite{margossian2019review}. 

In the next section, we apply the derived results in this case study to conduct numerical experiments.

\section{Numerical Experiments} \label{sec:experiments}
We conducted numerical experiments to evaluate the computational efficiency and safety of our proposed filter. \Cref{sec:exp-simulation-scenarios} describes the simulation environment and traffic scenarios. \Cref{sec:exp-computational-efficiency} demonstrates computational efficiency, and \cref{sec:exp-safety} validates safety. \Cref{sec:exp-limitations} discusses limitations of our approach. All simulations were conducted in Python on an Apple M2 Pro (16 GB RAM), using \texttt{CVXPY} \cite{diamond2016cvxpy} to solve the \ac{cbf}‐\ac{qp}. The code that reproduces our experimental results, along with a video demonstration, is publicly available\footnote{\href{https://github.com/bassamlab/SigmaRL}{https://github.com/bassamlab/SigmaRL}}.

\subsection{Simulation Environment and Traffic Scenarios} \label{sec:exp-simulation-scenarios}
\subsubsection{Simulation Environment}
We ran simulations in our \texttt{SigmaRL} \cite{xu2024sigmarl}, an open-source framework for motion planning of \acp{cav}. It provides various benchmarking traffic scenarios and also allows for rapid customization of traffic maps in the format of OpenStreetMap (OSM) \cite{haklay2008openstreetmap}. To challenge our safety filter, we trained an \emph{undertrained} \ac{rl} policy that could not reliably avoid collisions with road boundaries. For each traffic scenario, a vehicle was assigned a random predefined reference path and initialized at a random position and velocity along that reference path. Whenever the vehicle collided with a road boundary or reached the end of the reference path, we randomly reset its reference path, position, and velocity.

\subsubsection{Traffic Scenarios} \label{sec:exp-scenarios}
\begin{figure*}[t!]
    \centering
    \begin{subfigure}{0.24\linewidth}
        \centering
        \includegraphics[width=\linewidth]{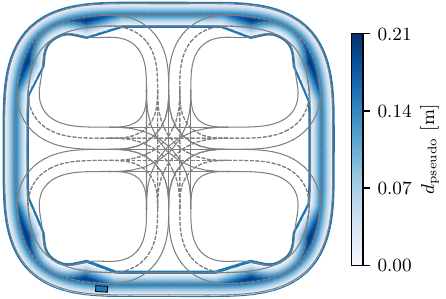}
        \caption{1st scenario.}\label{fig_pseudo_distance_CPM_entire_ref_0}
    \end{subfigure}
    \hfill
    \begin{subfigure}{0.17\linewidth}
        \centering
        \includegraphics[width=\linewidth]{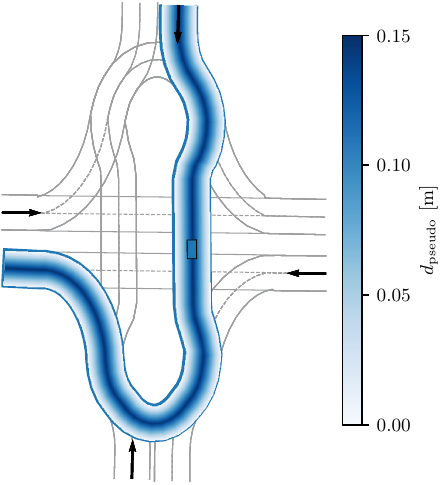}
        \caption{2nd scenario.} \label{fig_pseudo_distance_interchange_3_ref_7}
    \end{subfigure}
    \hfill
    \begin{subfigure}{0.17\linewidth}
        \centering
        \includegraphics[width=\linewidth]{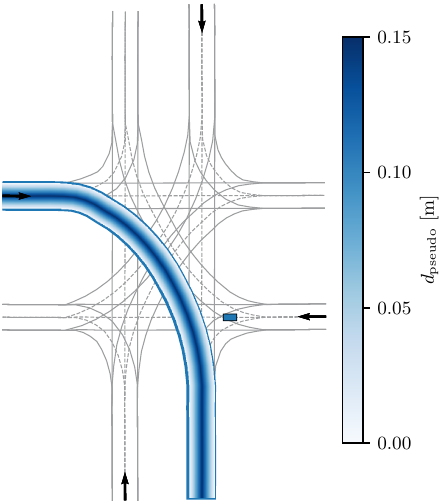}
        \caption{3rd scenario.} \label{fig_pseudo_distance_intersection_4_ref_8}
    \end{subfigure}
    \hfill
    \begin{subfigure}{0.23\linewidth}
        \centering
        \includegraphics[width=\linewidth]{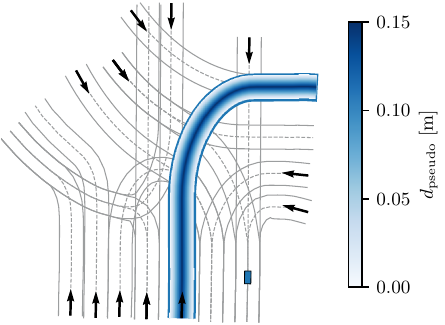}
        \caption{4th scenario.} \label{fig_pseudo_distance_intersection_6_ref_6}
    \end{subfigure}
    \caption{Visualization of the pseudo-distance of some selected roads of the four traffic scenarios.}
    \label{fig_pseudo_distance}
\end{figure*}

We evaluated four scenarios with complex road geometry, depicted in \cref{fig_pseudo_distance}. The 1st scenario uses the map of the \ac{cpmlab}, a small-scale testbed for \acp{cav} \cite{kloock2021cyberphysical}, featuring an eight‐lane intersection, a loop-shaped highway, and multiple highway merges. The 2nd scenario is a highway interchange near Waldkappel, Hesse, Germany (federal A44 to state B7\footnote{\url{https://www.openstreetmap.org/?mlat=51.145556&mlon=9.9225\#map=18/51.145674/9.922746}}). The 3rd scenario is a typical urban intersection. The 4th scenario is an artificial intersection with complex curvy lanes.

The areas in blue in \cref{fig_pseudo_distance} show the computed pseudo-distance introduced in \cref{sec:pseudo-distance} for some roads. The color represents the pseudo-distance from each point on the map to the road boundaries, calculated as the minimum of the pseudo-distances to the left and right road boundaries. The lane width in the 2nd, 3rd, and 4th scenarios is \SI{0.15}{\meter}, which results in a maximum pseudo-distance of \SI{0.15}{\meter} (all roads are two-lane in one direction). In the 1st scenario, the maximum pseudo-distance is higher (\SI{0.21}{\meter}) due to highway merge-ins and -outs, where the road boundaries are extended at the merging regions.

\subsubsection{Simulation Parameters} \label{sec:exp-parameters}
We model the vehicle footprint as a rectangle of length $\ell=\SI{0.16}{\meter}$ and width $\SI{0.08}{\meter}$. We approximate it with three circles of radius $r_{\mathrm{cir}}=\SI{0.048}{\meter}$ (computed from \eqref{eq:minimum-radius}), spaced \(d_{\mathrm{cir}}=\SI{0.053}{\meter}\) apart along the longitudinal axis.  We used a sampling period of $\Delta t=\SI{0.05}{\second}$.  For each traffic scenario, we ran five simulations with different random seeds, with each run lasting for \SI{30}{\second} (600 time steps). For the collision-avoidance constraints \eqref{eq:ttcbf-rereformulated}, we applied a linear class $\mathcal{K}$ function $\alpha(h^{j,i}) = \alpha_1 h^{j,i}$ with coefficient $\alpha_1=0.1$, and treated $\gamma \Delta t^{3} \approx 0$, given a sufficiently small $\Delta t$. The cost weighting matrix $R$ is diagonal and penalizes deviations from the nominal acceleration $u_{\mathrm{nom},v}$ more heavily than from the nominal steering rate $u_{\mathrm{nom},\delta}$. Consequently, the \ac{cbf}-\ac{qp} adjusts $u_{\mathrm{nom},\delta}$ more preferentially than $u_{\mathrm{nom},v}$. \Cref{tab:parameters} summarizes all parameters.

\begin{table}[t!]
    \caption{Simulation parameters.}
    \centering
    \begin{tabular}{ll}
        \toprule
        Parameter & Value \\
        \midrule
        Vehicle length $\ell$, width $w$ & \SI{0.16}{\meter}, \SI{0.08}{\meter} \\
        Vehicle wheelbase $\ell_{wb}$, rear wheelbase $\ell_{r}$ & \SI{0.16}{\meter}, \SI{0.08}{\meter} \\
        Number of circles $n_{\mathrm{cir}}$, radius $r_{\mathrm{cir}}$ & 3, \SI{0.048}{\meter} \\
        $\Delta t$, each simulation duration & \SI{0.05}{\second}, \SI{30}{\second} \\
        Class $\mathcal{K}$ function $\alpha_1$, $\gamma \Delta t^{3}$ in \eqref{eq:ttcbf-rereformulated} & 0.1, $\approx 0$ \\
        Weighting matrix $R$ in \eqref{eq:ocp-objective-function} & $\begin{bmatrix}
                                    30 & 0 \\
                                    0 & 1
                                    \end{bmatrix}$ \\
        Max. (min.) acceleration $u_{v,\text{max}}$ in \eqref{eq:ocp-u-max-min} & \SI{40}{\meter\per\second\squared} (\SI{-40}{\meter\per\second\squared}) \\
        Max. (min.) steering rate $u_{\delta,\text{max}}$ in \eqref{eq:ocp-u-max-min} & \SI{40}{\radian\per\second} (\SI{-40}{\radian\per\second}) \\
        \bottomrule
    \end{tabular}
    \label{tab:parameters}
\end{table}

\subsection{Computational Efficiency} \label{sec:exp-computational-efficiency}
In this section, we evaluate the real-time efficiency of our safety filter. Since the computation time strongly depends on the number of circles used to approximate the vehicle, we vary this number from one to five to assess its effect.

\begin{figure}[t!]
    \centering
    \includegraphics[width=1.0\linewidth]{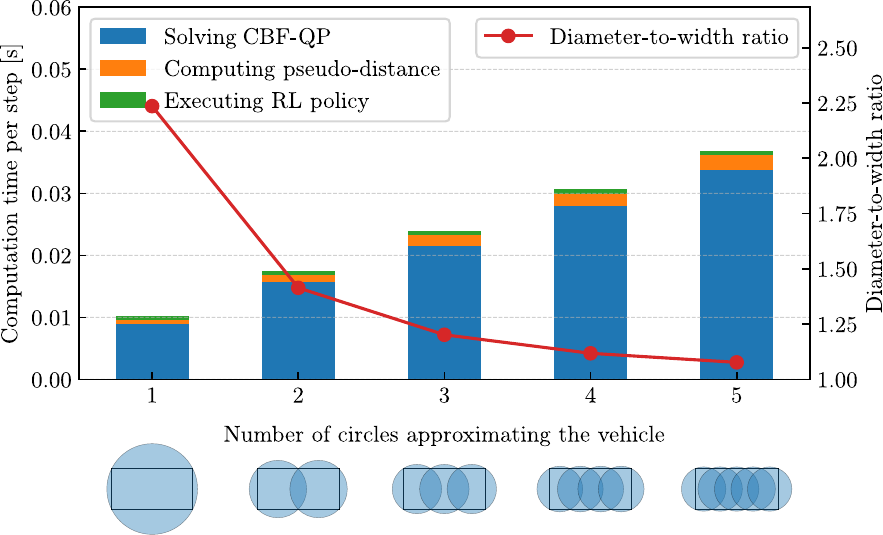}
    \caption{Computation times (solving \ac{cbf}-\ac{qp}, computing pseudo-distance, and executing \ac{rl} policy) and diameter-to-width ratio w.r.t. the number of circles approximating the vehicle.}
    \label{fig_computation_time}
\end{figure}

\Cref{fig_computation_time} depicts the computation time of solving the \ac{cbf}-\ac{qp} problem within our safety filter, computing the pseudo-distance, and executing the \ac{rl} policy. Among these three components, the execution time of the \ac{rl} policy is negligible, remaining below \SI{1.0}{\milli\second}. The computation times for both solving the \ac{cbf}-\ac{qp} and computing the pseudo-distance increase approximately linearly with the number of circles $n_{\mathrm{cir}}$. This is because both the number of collision-avoidance constraints and the number of calls of the pseudo-distance function grow linearly with $n_{\mathrm{cir}}$.  In practice, the number of circles should be chosen based on the vehicle geometry, considering the trade-off between the conservatism introduced by the approximation and the execution frequency of the safety filter. Moreover, computing the pseudo-distance remains efficient because one can limit the vehicle's scope to nearby path segments, and, importantly, the gradients and Hessians are inexpensive to compute due to vectorized computation. As shown by the red curve in \cref{fig_computation_time}, increasing the number of circles reduces the diameter-to-width ratio, which is the ratio between the diameter of each circle and the width of the vehicle. A lower diameter-to-width ratio implies a less conservative overapproximation of the vehicle geometry in the lateral direction. We do not consider the longitudinal direction, as the conservatism in that direction is significantly less dominant. In the simulations presented in the next section, we use three circles, resulting in a total computation time of less than \SI{25}{\milli\second} per step. This allows the safety filter to operate at a frequency of up to \SI{40}{\hertz} with this configuration. Note that using more circles can be unnecessary because it leads to a higher computation time while the reduction in conservatism saturates, as depicted by the red curve in \cref{fig_computation_time}.

\begin{figure*}[t!]
    \centering
    \begin{subfigure}{0.24\linewidth}
        \centering
        \includegraphics[width=\linewidth]{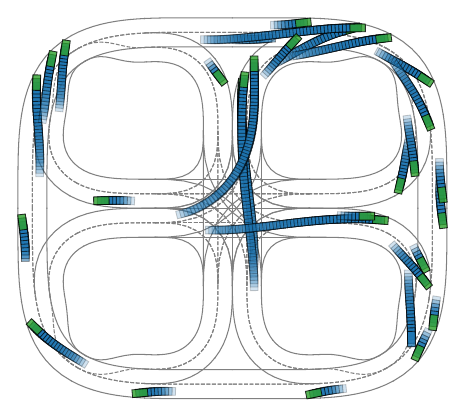}
        \caption{1st scenario: 29 collisions.}\label{fig_footprints_rl_seed_scenario_CPM_entire_1_circles_3}
    \end{subfigure}
    \hfill
    \begin{subfigure}{0.17\linewidth}
        \centering
        \includegraphics[width=\linewidth]{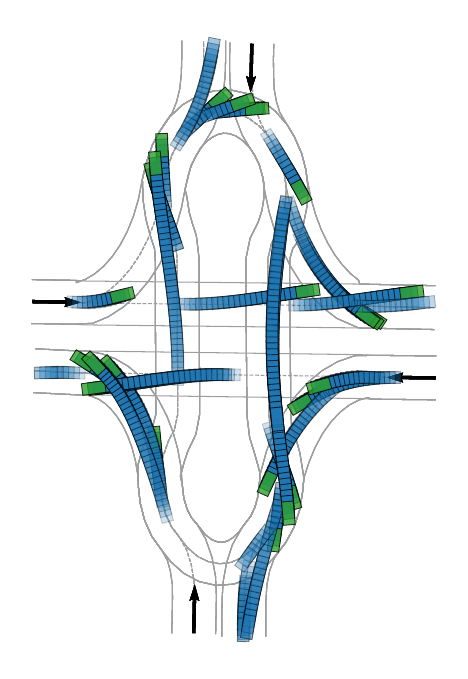}
        \caption{2nd scenario: 26.} \label{fig_footprints_rl_seed_scenario_interchange_3_1_circles_3}
    \end{subfigure}
    \hfill
    \begin{subfigure}{0.17\linewidth}
        \centering
        \includegraphics[width=\linewidth]{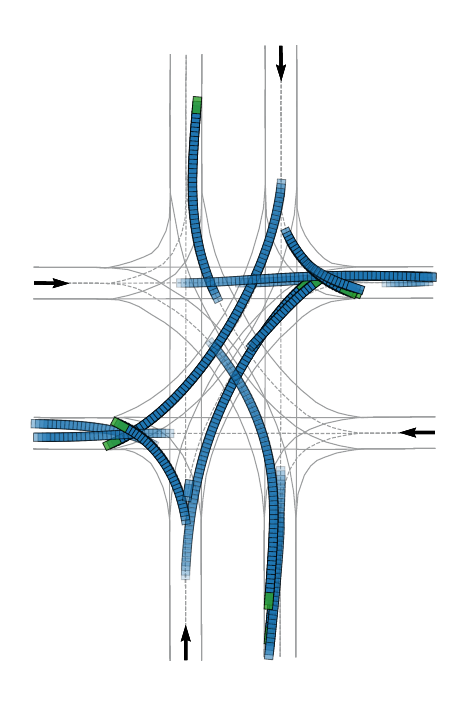}
        \caption{3rd scenario: 9.}\label{fig_footprints_rl_seed_scenario_intersection_4_1_circles_3}
    \end{subfigure}
    \hfill
    \begin{subfigure}{0.24\linewidth}
        \centering
        \includegraphics[width=\linewidth]{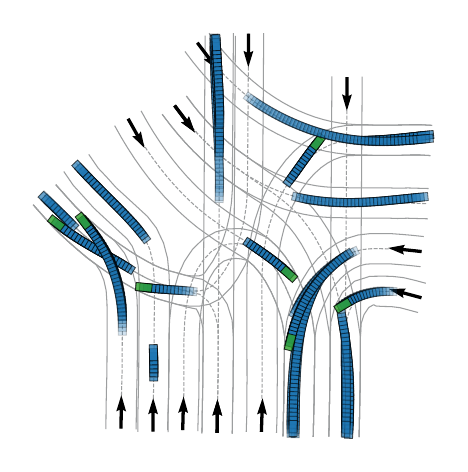}
        \caption{4th scenario: 8 collisions.}\label{fig_footprints_rl_seed_scenario_intersection_6_1_circles_3}
    \end{subfigure}
    \caption{Vehicle footprints with the undertrained \ac{rl} policy during 30-second simulations (600 time steps) in four traffic scenarios. Colliding footprints are shown in green. Black arrows indicate road entry directions.}
    \label{fig_footprints_rl}
\end{figure*}

\begin{figure*}[t!]
    \centering
    \begin{subfigure}{0.24\linewidth}
        \centering
        \includegraphics[width=\linewidth]{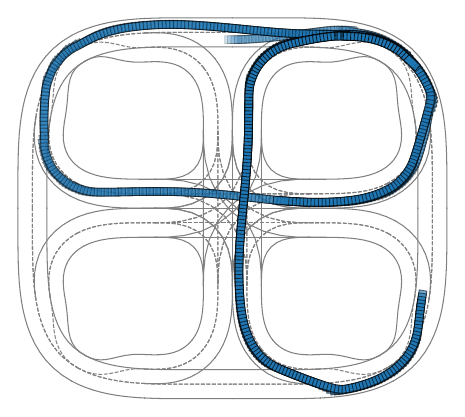}
        \caption{1st scenario: 0 collision.}\label{fig_footprints_rl_cbf_seed_scenario_CPM_entire_1_circles_3}
    \end{subfigure}
    \hfill
    \begin{subfigure}{0.17\linewidth}
        \centering
        \includegraphics[width=\linewidth]{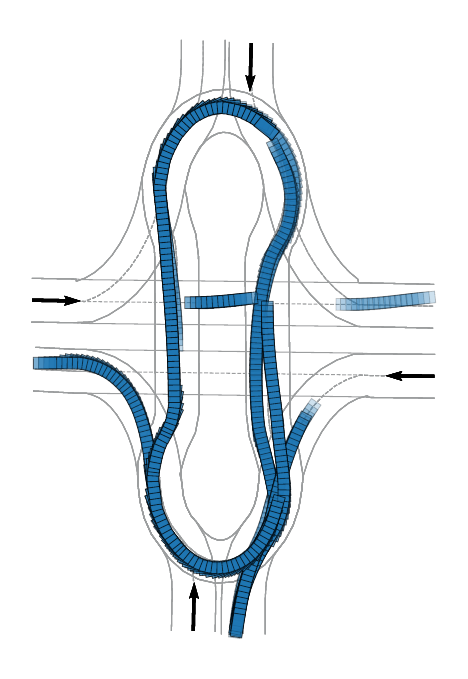}
        \caption{2nd scenario: 0.} \label{fig_footprints_rl_cbf_seed_scenario_interchange_3_1_circles_3}
    \end{subfigure}
    \hfill
    \begin{subfigure}{0.17\linewidth}
        \centering
        \includegraphics[width=\linewidth]{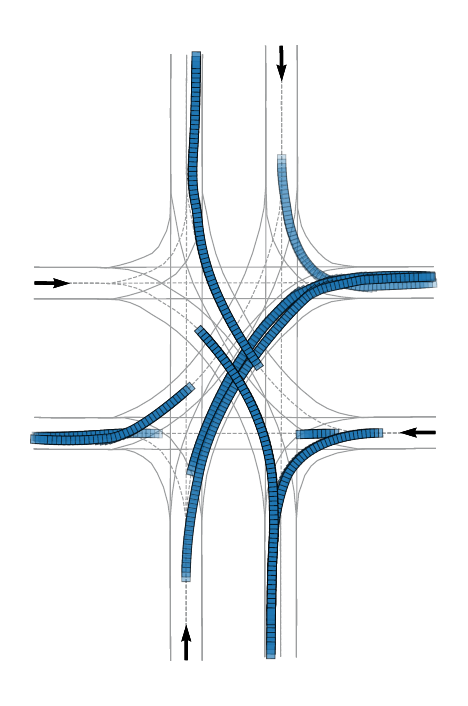}
        \caption{3rd scenario: 0.}\label{fig_footprints_rl_cbf_seed_scenario_intersection_4_1_circles_3}
    \end{subfigure}
    \hfill
    \begin{subfigure}{0.24\linewidth}
        \centering
        \includegraphics[width=\linewidth]{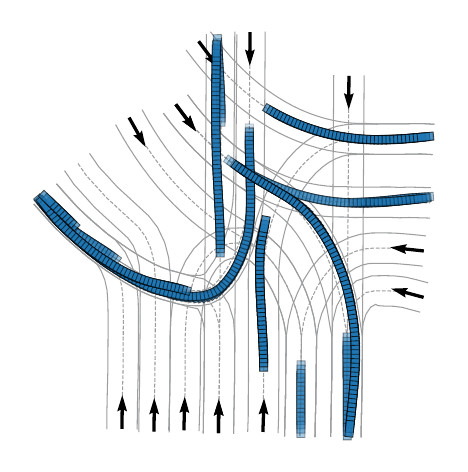}
        \caption{4th scenario: 0 collision.}\label{fig_footprints_rl_cbf_seed_scenario_intersection_6_1_circles_3}
    \end{subfigure}
    \caption{Vehicle footprints with the undertrained \ac{rl} policy certified by our safety filter during 30-second simulations (600 time steps) in four traffic scenarios. Colliding footprints are shown in green. Black arrows indicate road entry directions.}
    \label{fig_footprints_rl_cbf}
\end{figure*}

\subsection{Safety} \label{sec:exp-safety}
In this section, we demonstrate the safety of our safety filter in the four traffic scenarios introduced in \cref{sec:exp-scenarios}. We use $n_{\mathrm{cir}} = 3$ circles to approximate the vehicle. Recall that we conducted five runs with different random seeds for each traffic scenario. Each run lasted for 30 seconds, corresponding to 600 time steps.

\Cref{fig_footprints_rl} shows the footprints of the vehicle controlled by the undertrained \ac{rl} policy (without our safety filter) at all time steps during one of the five runs for each traffic scenario. The colliding footprints are visualized in green. As shown, the vehicle frequently collided with the road boundaries, with 29, 26, 9, and 8 collisions in the four traffic scenarios, respectively. After applying our safety filter, no collisions occurred in any scenario, as shown in \cref{fig_footprints_rl_cbf}, which contains no green footprints. Since we conducted five runs per scenario, we averaged the number of collisions for each scenario. On average, the undertrained \ac{rl} policy caused 28.6, 26.2, 11.0, and 7.0 collisions in the four traffic scenarios, respectively, while our safety filter successfully avoided all collisions in all scenarios.

\begin{figure*}[t!]
    \centering
    \begin{subfigure}{0.24\linewidth}
        \centering
        \includegraphics[width=\linewidth]{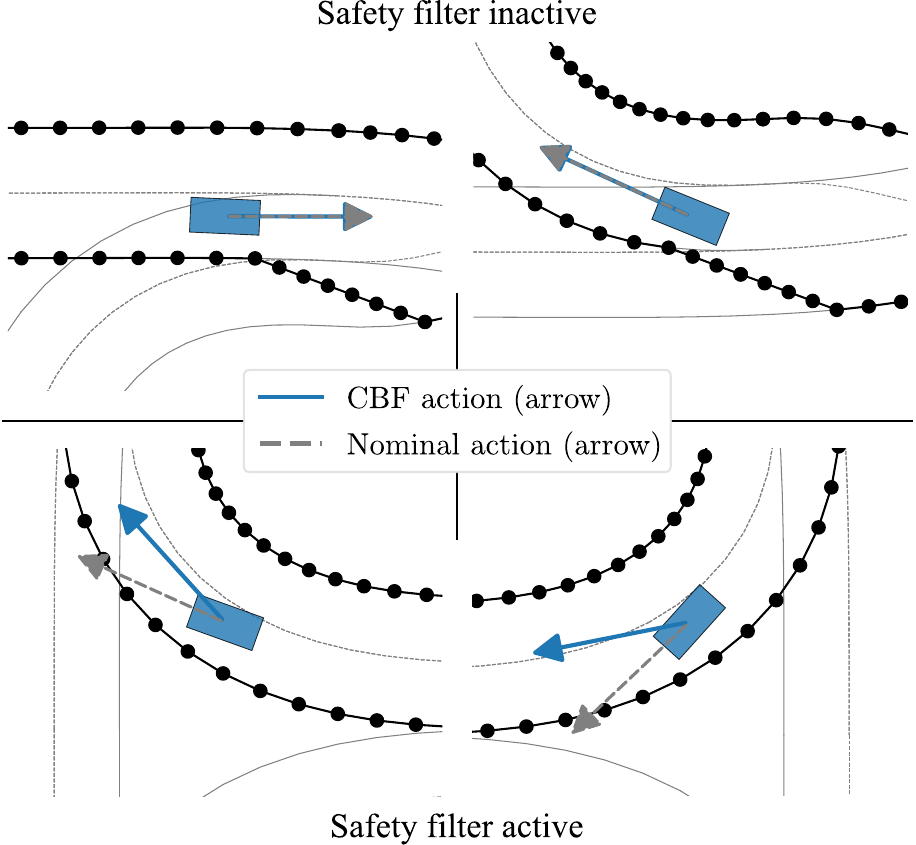}
        \caption{1st scenario.}\label{fig_cbf_action_rl_cbf_seed_scenario_CPM_entire_1_circles_3}
    \end{subfigure}
    \hfill
    \begin{subfigure}{0.24\linewidth}
        \centering
        \includegraphics[width=\linewidth]{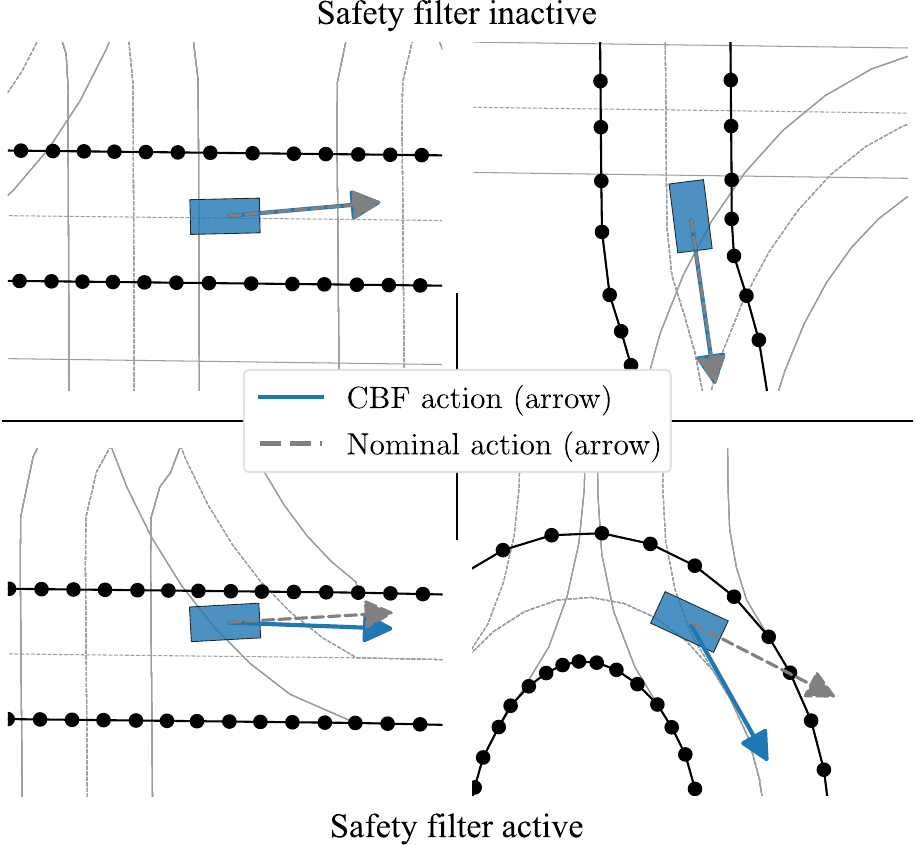}
        \caption{2nd scenario.} \label{fig_cbf_action_rl_cbf_seed_scenario_interchange_3_1_circles_3}
    \end{subfigure}
    \hfill
    \begin{subfigure}{0.24\linewidth}
        \centering
        \includegraphics[width=\linewidth]{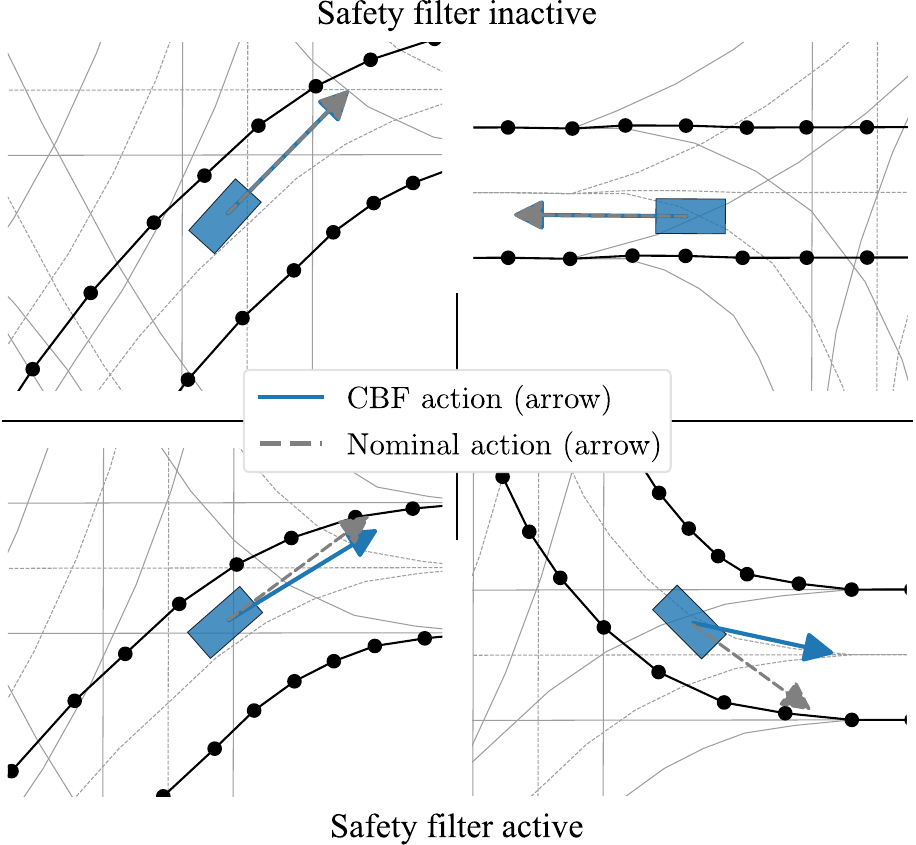}
        \caption{3rd scenario.}\label{fig_cbf_action_rl_cbf_seed_scenario_intersection_4_1_circles_3}
    \end{subfigure}
    \hfill
    \begin{subfigure}{0.24\linewidth}
        \centering
        \includegraphics[width=\linewidth]{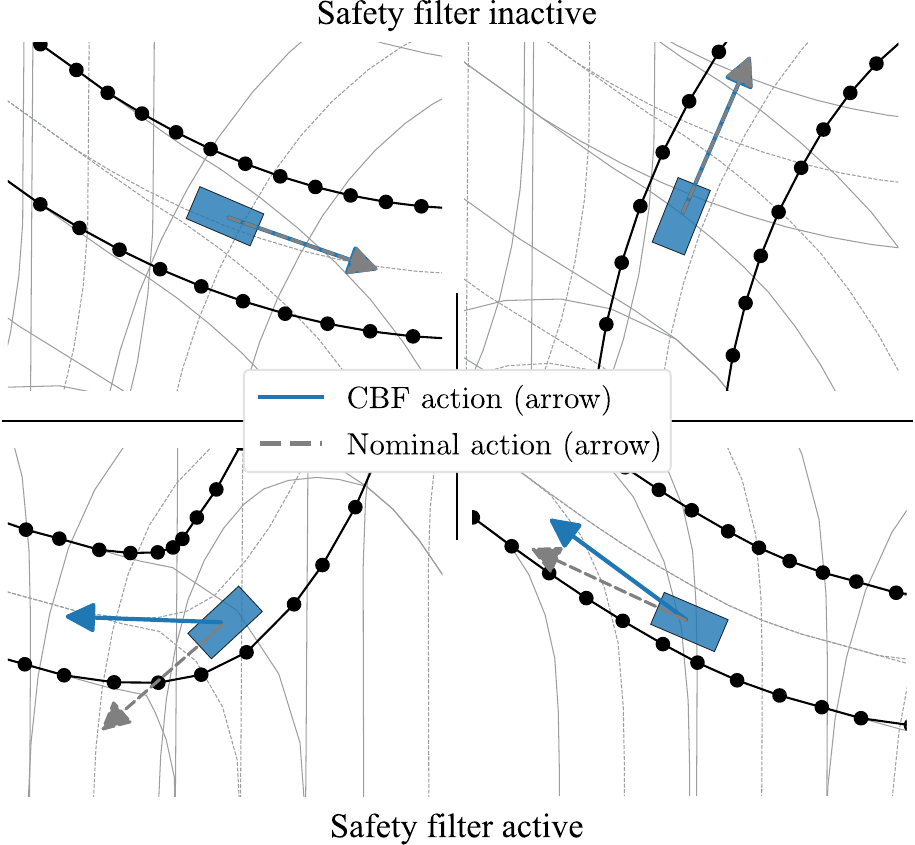}
        \caption{4th scenario.}\label{fig_cbf_action_rl_cbf_seed_scenario_intersection_6_1_circles_3}
    \end{subfigure}
    \caption{Selected snippets illustrating when our safety filter was inactive or active. Gray dashed arrows: nominal actions; blue solid arrows: \ac{cbf} actions. The filter was active when the two arrows in each snippet differed.}
    \label{fig_action_rl_cbf}
\end{figure*}

To further analyze our safety filter, we provide \cref{fig_action_rl_cbf}, which shows the nominal actions (gray dashed arrows) and the certified actions (called \ac{cbf} actions, blue solid arrows). If the two arrows overlap, the safety filter was inactive because the nominal action was safe. If they differ, the filter was active because the nominal action was unsafe. The polylines representing the road boundaries of the vehicle at the corresponding time step are shown as black dots, where the dots are the discrete points of the polyline. In \cref{fig_action_rl_cbf}, for each traffic scenario, we selectively provide four snippets corresponding to four time steps: two where the safety filter was inactive (upper) and two where it was active (bottom). Consider the first active case in \cref{fig_cbf_action_rl_cbf_seed_scenario_CPM_entire_1_circles_3} as an example. The nominal action points to the top left, which would move the vehicle towards the left boundary. The safety filter modified this to point more upward to prevent the vehicle from getting too close to the boundary. Similar behavior can be observed in the other active cases shown in \cref{fig_action_rl_cbf}.

\subsection{Discussions and Limitations} \label{sec:exp-limitations}
Although the pseudo-distance enables smooth and differentiable distance computations, real-world road boundaries can exhibit significant nonsmoothness, for example, due to abrupt lane narrowings or sudden detours caused by obstacles. These discontinuities can cause abrupt changes in the drivable set, potentially rendering the \ac{cbf}-\ac{qp} infeasible, especially under tight control bounds. One possible solution is to redesign the pseudo-distance to better capture such abrupt changes in road geometry, thereby yielding smoother distance computations. Another solution is to enhance the feasibility guarantees of the \ac{cbf}-\ac{qp} itself, which remains a fundamental challenge in \acp{cbf} \cite{xiao2022sufficient}.

Moreover, the current safety filter only considers collision avoidance with road boundaries and does not account for surrounding vehicles. This makes it particularly suitable for single-vehicle scenarios such as single-vehicle autonomous racing. However, since other vehicles can also be represented as polylines (closed polylines), it is possible to extend our approach to handle multi-vehicle scenarios.

\section{Conclusions}\label{sec:conclusions}
We proposed a real-time \ac{cbf}-based safety filter for safety certification of motion planning against collisions with arbitrarily shaped road boundaries represented as polylines. Unlike methods that rely on a conservative overapproximation of road boundaries, our safety filter considers their actual geometric shapes. In a case study, we demonstrated how to apply it to the nonlinear kinematic bicycle model. We conducted extensive numerical experiments in four traffic scenarios with complex roads. In each scenario, the safety filter was used to certify an undertrained \ac{rl} policy. While the \ac{rl} policy resulted in many collisions across the scenarios, our safety filter successfully prevented all of them. Furthermore, our safety filter executed in less than \SI{25}{\milli\second} per step, corresponding to an execution frequency of \SI{40}{\hertz}. These results indicate that it can serve as a practical safety filter for motion planning, such as learning-based methods that lack safety guarantees.

\bibliographystyle{IEEEtran}
\bibliography{00_literature}

\end{document}